\documentclass[10pt,twocolumn,letterpaper]{article}

\usepackage{iccv}              

\usepackage[accsupp]{axessibility}  

\definecolor{iccvblue}{rgb}{0.21,0.49,0.74}
\usepackage[pagebackref,breaklinks,colorlinks,allcolors=iccvblue]{hyperref}

\def\paperID{9} 
\def\confName{ICCV}
\def\confYear{2025}

\title{SynapFlow: A Modular Framework Towards Large-Scale Analysis of Dendritic Spines}

\author{Pamela Osuna-Vargas$^{1,2}$
\and
Altug Kamacioglu$^{3}$
\and
Dominik F. Aschauer$^{3}$
\and
Petros E. Vlachos$^{1}$
\and
Sercan Alipek$^{1,4}$
\and
Jochen Triesch$^{1,2}$
\and
Simon Rumpel$^{3}$
\and
Matthias Kaschube$^{1,2}$\\
\and
$^{1}$Frankfurt Institute for Advanced Studies, Frankfurt, Germany
\and
$^{2}$Institute of Computer Science, Goethe University Frankfurt, Frankfurt, Germany
\and
$^{3}$Institute of Physiology, Focus Program Translational Neurosciences, University Medical Center, \and Johannes Gutenberg University-Mainz, Mainz, Germany
\and
$^{4}$Mechanical Engineering Department, Universität Siegen
\and
{\tt\small \{osuna, pvlachos, triesch, kaschube\}@fias.uni-frankfurt.de}\\
{\tt\small \{akamacio, daschauer, sirumpel\}@uni-mainz.de}, {\tt\small \{sercan.alipek\}@uni-siegen.de}
}

\begin{document}
\maketitle
\begin{abstract}
Dendritic spines are key structural components of excitatory synapses in the brain. Given the size of dendritic spines provides a proxy for synaptic efficacy, their detection and tracking across time is important for studies of the neural basis of learning and memory. Despite their relevance, large-scale analyses of the structural dynamics of dendritic spines in 3D+time microscopy data remain challenging and labor-intense. Here, we present a modular machine learning-based pipeline designed to automate the detection, time-tracking, and feature extraction of dendritic spines in volumes chronically recorded with two-photon microscopy. Our approach tackles the challenges posed by biological data by combining a transformer-based detection module, a depth-tracking component that integrates spatial features, a time-tracking module to associate 3D spines across time by leveraging spatial consistency, and a feature extraction unit that quantifies biologically relevant spine properties.
We validate our method on open-source labeled spine data, and on two complementary annotated datasets that we publish alongside this work: one for detection and depth-tracking, and one for time-tracking, which, to the best of our knowledge, is the first data of this kind. To encourage future research, we release our data, code, and pre-trained weights at \url{https://github.com/pamelaosuna/SynapFlow}, establishing a baseline for scalable, end-to-end analysis of dendritic spine dynamics.
\end{abstract} 
\section{Introduction} 
\label{sec:introduction}
Synapses, the connections between brain cells, are highly dynamic structures whose changes tightly relate to memory and learning processes \cite{yang2009stably, kasai2010structural, holler2021structure, knott2006spine}, occurring even without external stimuli \cite{hazan2020activity, loewenstein2015predicting}. Most excitatory cortical connections terminate at dendritic spines, small protrusions from dendrites with bulbous heads connected via narrow necks. Since spine size correlates with synaptic efficacy \cite{matsuzaki2001dendritic, engert1999dendritic, kopec2006glutamate}, monitoring large numbers of spines could help reveal the neural underpinnings of natural intelligence, specifically the brain's capacity for plasticity and information storage \cite{Heck2023, Ma2022, murakoshi2012postsynaptic}. However, the study of spine dynamics has long been hampered by technical challenges in both imaging and analysis. While improved microscopy techniques \cite{meng2019high, birkner2017improved} now enable large-scale brain imaging, manual annotation remains impractical, particularly for chronic experiments, where manual analysis involves not only identifying 2D structures and resolving them across depth, but also tracking individual spines over time.

Methods for object detection in images have seen considerable progress since large-scale datasets such as ImageNet \cite{deng2009imagenet} and COCO \cite{lin2014microsoft} catalyzed the development of deep learning-based methods. A major breakthrough came with DETR \cite{carion2020end}, the first model to frame object detection as a direct set prediction problem using a transformer architecture. While innovative, DETR suffers from slow convergence and struggles with the detection of small objects, limitations largely attributed to its use of global attention over high-resolution features. To overcome these challenges, Deformable DETR \cite{zhu2020deformable} introduced deformable attention modules that focus on a sparse set of relevant features. Deformable DETR features an improved detection performance for small objects and an accelerated convergence.

Building on these advances in detection, object tracking has also evolved considerably. Single-object tracking focuses on following a single target through an image sequence, typically assuming its presence in all frames. Multi-object tracking (MOT) extends the challenge to track the identities of multiple objects that may appear, disappear, or occlude each other, making it a more challenging task. Progress here has also been driven by better detectors, especially in tracking-by-detection methods. End-to-end approaches like Trackformer \cite{meinhardt2022trackformer} unify detection and association, but often assume structured or predictable motion patterns \cite{wojke2017simple, meinhardt2022trackformer, bewley2016simple}. This assumption typically does not hold for longitudinal brain imaging data, where objects undergo complex, non-rigid local deformations across depth and imaging days. These deformations can arise from animal movement (e.g. breathing), slight changes in the imaging axis perspective (e.g. angle difference), and even imaging artifacts.

In this work, we propose a modular pipeline tailored for tracking dendritic spines in 3D+time microscopy data. Our framework consists of: a transformer-based detection module optimized for dendritic spines, a depth-tracking module that relies on spatial cues, a time-tracking module for linking 3D spine objects across time points, and a feature extraction module to quantify key properties of spines.

Moreover, we provide two novel, manually labeled, complementary datasets to foster progress in automated spine analysis: the first comprises 2D spine detections linked across depth, the second set contains spines tracked across imaging days, enabling objective evaluations of the modules of our analysis pipeline and of future methods.

Our contributions are as follows: 
(i) we train a Deformable DETR model on a robust and diverse dataset to detect dendritic spines, achieving strong performance compared to existing methods without added computational cost;
(ii) we propose two simple yet effective methods for reliable tracking across depth, based on spatial and visual similarity;
(iii) we introduce a strategy for time-tracking dendritic spines;
(iv) we develop a module to extract biologically relevant features, including size and spine-to-dendrite distance; (v) we release two novel datasets: one for 2D spine detection and depth-tracking, and a second one for time-tracking of dendritic spines, the first publicly available dataset of its kind. 

We evaluate each module using ours and different open-source manually annotated datasets, demonstrating generalizability across experimental settings. Together, the four modules and the accompanying data form an integrated framework that establishes a baseline for large-scale, end-to-end analysis of dendritic spines. Ultimately, this approach supports the systematic study of spine dynamics over space and time, contributing to a deeper understanding of their role in memory and learning in the brain. 

\subsection{Related work}
One still common approach is to perform a manual annotation of a limited number of spines before proceeding to their morphological analysis \cite{basu2018quantitative, pchitskaya2023spinetool, kashiwagi2019computational, XIAO201825}.
Recent methods for automated dendritic spine analysis have primarily focused on detection. SpineS, introduced by Argunşah et al. \cite{arguncsah2022interactive}, identifies regions of interest (ROIs) using Speeded-up Robust Features (SURF), followed by CNN-based classification. Fernholz et al. \cite{fernholz2024deepd3} propose a segmentation approach in which 2D images are encoded into a shared latent space, and decoded into separate spine and dendrite masks. Individual 3D spines are then reconstructed through connected component analysis. Vogel et al. \cite{vogel2023utilizing} use a CNN-based detector, achieving the best performance with Faster R-CNN and Cascade R-CNN, and further use the Intersection over Minimum (IoM) to group 2D instances into 3D spine objects. Building on this approach, Eggl et al. \cite{eggl2024spyden} train a Faster-RCNN model to detect bounding boxes and estimate spine centers, which are further extended into polygonal shapes using a ray-based heuristic. Despite its user-friendly graphical interface, this method requires manual input to trace the dendritic skeleton, limiting full automation. Although these methods represent valuable advances in spine detection and have been instrumental for neuroscience research, the subsequent steps needed for the analysis of time-lapse 3D data remain largely unaddressed. 

A limited number of open-access dendritic spine datasets are currently available. Amongst the most widely used are the dataset provided by Smirnov-Garrett \cite{tavita2018labeled}, and more recently the DeepD3 dataset \cite{fernholz2023data} and the dataset provided in Vogel et al. \cite{vogel2023utilizing}. These datasets include raw microscopy images with annotations identifying dendritic spines. The datasets from Smirnov-Garrett and Vogel et al. provide annotations in the form of 2D bounding boxes marking individual spines, whereas the DeepD3 dataset offers two types of labels: binary segmentation masks for dendrites and for spines, and 2D coordinates indicating the center of mass of each 2D spine instance.
While valuable, these datasets are limited to static 2D annotations and do not include temporal or volumetric tracking information, which are essential for the investigation of structural changes in spine size and morphology.
\section{Methods}
\label{sec:method}

\subsection{Pipeline overview}
\begin{figure}[hbt!]
  \centering
  \includegraphics[width=0.5\textwidth]{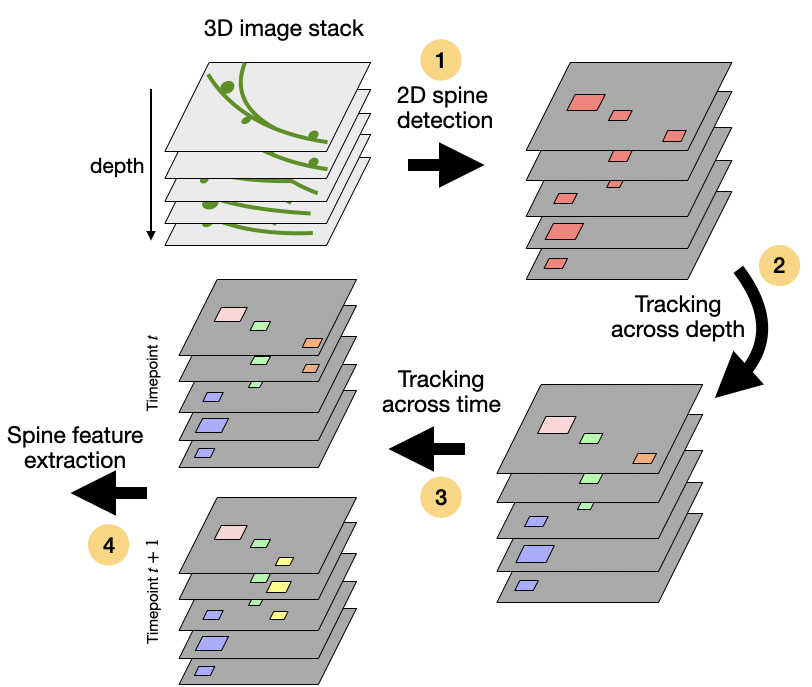}
  \caption{\textbf{Overview of our modular framework to reveal dynamic changes of dendritic spines.} \textbf{1}: 2D slices from 3D image volumes are fed into the detection model. \textbf{2}: 2D detections from the same volume are matched across depth to reveal individual 3D spine objects. \textbf{3}: Having built 3D spine objects in two volumes of the same field of view imaged at different timepoints, identities are resolved over time. \textbf{4}: Spine features are extracted and monitored across time. Here, we focus on spine size (a proxy for synaptic efficacy) and the distance from the dendrite to the spine head.}
  \label{fig: pipeline_overview}
\end{figure}

Our proposed framework SynapFlow is composed of individual modules that enable dendritic spine detection, tracking across depth to reconstruct 3D objects, tracking across time, and the extraction of relevant information from the objects, such as spine size, and spine-to-dendrite distance. An overview of our framework is illustrated in Fig. \ref{fig: pipeline_overview}. The following sections describe the datasets and the approach used at every step of the pipeline. The training process and implementation details can be found in the Supplement.

\subsection{Datasets}
\label{subsec:dataset}
The labeled datasets employed to develop and validate our pipeline are based on 3D image volumes (stacks) acquired using chronic in vivo two-photon microscopy from the auditory cortex of GFP-M transgenic mice as described in \cite{vogel2023utilizing}. Each voxel measures approximately $0.1075$ \textmu m $\times 0.1075$ \textmu m $\times 0.5$ \textmu m, thus slicing the same spine, typically, at multiple depths, and each image slice has a size of $512\times 512$ pixels.

From these data we generate two manually-labeled data sets, published here alongside our analysis pipeline. The first one, referred to as S3D, is used for training and validating (using different image volumes) the detection and depth-tracking modules of our pipeline (see Supplement for a detailed breakdown of its content). Based on the annotations in  \cite{vogel2023utilizing}, we re-annotated the data to obtain consistent depth-wise identity assignments, and to tighten the bounding boxes, avoiding the inclusion of the dendritic region, often overestimated in the original labels. These updated annotations consist of 2D bounding boxes and the 3D spine identity. Spines located directly on top of or below the dendrite were deliberately excluded due to their ambiguous morphology and proneness to size miscalculation. 

The second manually-labeled data set, referred to as S2D+T, is used to validate the time-tracking method and the estimation of spine size and morphology. An overview of its content is reported in the Supplement. Spines were tracked across four timepoints: t, t + 2 hours, t + 3 days, and t + 3 days 2 hours. Spines labeled were chosen from a specific subregion of the image that was visible at all timepoints, and only the most representative 2D instance was labeled for every 3D object. Therefore, this subset contains partial 2D detections tracked across four time steps. One 2D instance is annotated as a polygon that includes spine head and neck. Spine size, and spine-to-dendrite length are manually estimated as described in \cite{loewenstein2011multiplicative}.

In addition to our own dataset, we also evaluate our detection model on two open-source datasets. The first is the DeepD3 dataset, published alongside the DeepD3 framework \cite{fernholz2023data, fernholz2024deepd3} (the evaluation, but not the training set was made publicly available by the authors). The DeepD3 dataset consists of a volume of size $71 \times 366 \times 1444$ pixels, imaged from a brain slice of the rat hippocampus, and shows densely populated dendrites. The second dataset, from Smirnov-Garrett \cite{tavita2018labeled}, contains small images of variable size, ranging from $128 \times 128$ pixels to $244 \times 212$ pixels, of the same field of view imaged multiple times (although no information about their timestamp is given) from mouse organotypic hippocampal cell cultures (see Supplement for an overview of the datasets).

The DeepD3 dataset provides two types of ground-truth labels for each image slice in the volume: (i) 2D coordinates marking the center of each 2D spine, and (ii) binary masks separating spine pixels from non-spine pixels. Since both formats differ from ours (2D bounding boxes), we first select the binary mask annotations and convert them to match our format. Specifically, we first compute the intersection of the masks labeled by their three annotators, then apply connected component analysis to each slice (following its implementation in the DeepD3 GUI) to separate individual 2D spines. Each segmented component is then approximated with a bounding box, which we use as the ground-truth for the DeepD3 dataset.

\subsection{Dendritic spine detection in 2D}
\label{subsec: method-detection}
Expanding on our previous work \cite{vogel2023utilizing}, we train a Deformable DETR network with two-stage refinement \cite{zhu2020deformable} pre-trained on the COCO dataset. A ResNet50 is used as the CNN backbone.

\subsection{Spine tracking across depth}
\label{subsec: method-depth-tracking}
Using the S3D dataset, a Siamese neural network $f_{\theta}$ is trained to encode spine image patches into embeddings, such that these are similar if the patches belong to the same 3D object, and dissimilar otherwise. This is achieved by learning to minimize a contrastive loss function 
\begin{equation}
\label{eq: loss-depth-tracking}
\begin{aligned}
    \mathcal{L}_\text{cont}(\mathbf{x}_i, \mathbf{x}_j, \theta) = 
    \mspace{370mu}
    \\
    \mathbf{1}_{[y_i = y_j]} \left \|f_{\theta}(\mathbf{x}_i) - f_{\theta}(\mathbf{x}_j) \right \| _{2}^2 
    \mspace{240mu}
    \\
    + \mathbf{1}_{[y_i \neq y_j]}\max\big (0, \epsilon - \left \| f_{\theta}(\mathbf{x}_i)-f_{\theta}(\mathbf{x}_j)\right \|_{2}^2\big)^2
    \mspace{90mu}
    \\
\end{aligned}
\end{equation}
for image pairs $(\mathbf{x}_i, \mathbf{x}_j)$, with their 3D identity labels $(y_i, y_j)$, respectively, where $\epsilon$ controls the maximal possible loss for instances with different 3D identities.

Once trained, we use the network $f$ to encode image patches into embeddings $\mathbf{v}_i$ and $\mathbf{v}_j$, where $f(\mathbf{x}_k) = \mathbf{v}_k$ for $k = i, j$. We then calculate the Euclidean distance between these embeddings to obtain the appearance cost $C_\text{app}$, which estimates the visual dissimilarity between spine patches. Beyond appearance, we also estimate spatial consistency by computing the generalized intersection over union (gIoU) \cite{rezatofighi2019generalized} between the bounding boxes (see Supplement for a schematic of the spatial and appearance costs). The total cost is therefore the weighted sum

\begin{equation}
\label{eq: cost_depth_tracking}
\begin{aligned}
    C_\text{MatchInDepth}(i, j) = 
    \mspace{220mu}
    \\
    \lambda_\text{app} C_\text{app}(\mathbf{v}_i, \mathbf{v}_j) 
    + \lambda_\text{sp} C_\text{sp}(\mathbf{b}_i, \mathbf{b}_j),
    \mspace{10mu} \\
\end{aligned}
\end{equation}
where $\lambda_\text{app}$ and $\lambda_\text{sp}$ are the weighting factors, empirically set. 

After building a cost matrix of the potential matches, we find the optimal identity association across depth using the Hungarian algorithm, and discard pairs whose matching cost is higher than a given threshold, here set to 0.5.

\subsection{Spine tracking across time}
We track 3D spine objects across consecutive imaging timepoints by defining a matching cost that combines spatial consistency, appearance similarity, and depth consistency.

The appearance cost $C_\text{app}$ measures the dissimilarity between appearance embeddings at consecutive time points $t$ and $t+1$. We encode the image patches into embeddings using the same network trained for depth tracking (Section \ref{subsec: method-depth-tracking}). For each 3D spine object $i$, we compute the mean appearance embedding $\bar{\mathbf{v}}_i^t$ from its 2D instances at time $t$, and analogously for time $t+1$. $C_\text{app}$ is calculated as the Euclidean distance between these mean embeddings.

The spatial consistency cost $C_{\text{sp}}$ accounts for expected spatial relationships between spines across time. We first compute a maximum intensity projection (MIP) of each 3D volume, reducing it to a 2D representation. We then estimate dense pixel-level correspondences between consecutive MIPs using Pump \cite{revaud2022pump}. Using the resulting deformation field, we project the bounding box coordinates from one timepoint onto the next, and compute the gIoU between the projected and actual bounding boxes. Here, $\text{med}(B_i^{t})$ denotes the median boundig box coordinates of all 2D detections belonging to 3D spine object $i$ at time $t$.

The depth consistency cost $C_\text{depth}$ measures overlap between their respective depth layers occupied by spine objects at consecutive timepoints. After aligning 3D volumes along the z-axis using cross-correlation, we compare the discrete sets of depth layers $Z^{i}$, $Z^{j} \subset \mathbb{Z}$ associated with each spine object. Then we define

\begin{equation}
\label{eq: z-cost}
\begin{aligned}
C_\text{depth}(Z^i, Z^j) =
\hspace{5.6cm}  &\\
\begin{cases} 
      \max\big(Z_{\min}^i - Z^{j}_{\max}, Z^{j}_{\min} - Z^k_{\max}\big), & if Z^i \cap Z^{j} = \varnothing \\
      1- \dfrac{|Z^i \cap Z^{j}|}{\min\big(|Z^i|, |Z^{j}|\big)}, & \text{otherwise} \\
   \end{cases}
   \end{aligned}
\end{equation}
where $Z^i_{\min} = \min Z^i$, $Z^i_{\max} = \max Z^i$, $Z^{j}_{\min} = \min Z^{j}$, and $Z^{j}_{\max} = \max Z^{j}$.

The total cost for associating spine objects $i$ and $j$ across timepoints $t$ and $t+1$ is:
\begin{equation}
\label{eq: cost-time-tracking}
\begin{aligned}
    C_\text{MatchInTime}(i, j) = 
    \lambda_\text{sp} C_\text{sp}(\text{med}(B_i^{t}), \text{med}(B_j^{t+1}))
    \mspace{100mu}
    \\
     + \lambda_\text{depth} C_\text{depth}(Z_i^{t}, Z_j^{t+1}) + \lambda_\text{app} C_\text{app}(\bar{\mathbf{v}}^{t}_i, \bar{\mathbf{v}}^{t+1}_j),
    \mspace{35mu}
    \\
 \end{aligned}
\end{equation}
where $\lambda_{\text{sp}}$, $\lambda_{\text{depth}}$ and $\lambda_{\text{app}}$ are scalar weights controlling the relative influence of each term.

\subsection{Computation of spine features}
\label{subse:method-feature-extraction}
To estimate spine sizes, we automatize the manual procedure in Loewenstein et al. \cite{loewenstein2011multiplicative}, where the spine volume is approximated by its integrated fluorescence intensity in 2D, normalized by the dendrite intensity (see Supplement for implementation details).

To estimate the spine-to-dendrite distance, we first dilate the spine surface so that the spine–dendrite junction can be approximated as the region where the dilated spine overlaps with the dendrite (excluding background and other spines). We then compute the center of the spine head and measure the Euclidean distance between these two points (see implementation details in the Supplement).

\subsection{Performance evaluation} 
\label{subsec:method-performance-eval}
Detection accuracy alone is assessed using the S3D dataset and standard metrics, precision, recall, and F1-score. 

For depth-tracking evaluation, also based on the S3D dataset, the task is reformulated as a multi-object tracking (MOT) problem in 2D, where spine objects are linked across consecutive depth slices. We evaluate performance using standard MOT metrics implemented in \textit{TrackEval} \cite{luiten2020trackeval}: Multiple Order Tracking Accuracy (MOTA) \cite{bernardin2008evaluating}, which penalizes missed detections, false positive detections, and identity switches; Higher Order Tracking Accuracy (HOTA) \cite{luiten2021hota}, which balances detection, association, and localization performance in a unified score; Identification F1-score (IDF1) \cite{ristani2016performance}, which measures the consistency of predicted trajectories by computing the F1-score over matched identities, and Association Accuracy (AssA) \cite{luiten2021hota}, which isolates the quality of object identity preservation. It is important to note that the emphasis on detection versus tracking accuracy varies across MOT metrics: while MOTA prioritizes detection, HOTA balances detection and identity, IDF1 emphasizes identity consistency, and AssA focuses almost entirely on association quality.

We evaluate time-tracking using the S2D+T dataset, which provides sparse ground-truth annotations (one 2D representative per 3D spine object) with their temporal identities for a subset of spine objects. We assess our time-tracking method by taking ground-truth detections and evaluating how accurately our automated approach assigns consistent identities across timepoints. This isolates the performance of the temporal association component by comparing predicted identity assignments to ground-truth tracks using the same MOT metrics described above.
Spine feature extraction is evaluated by computing the correlation between the manually measured and automatically estimated features. This assessment is done under two conditions: using ground-truth detections to isolate feature computation accuracy, and using automated detection and depth-tracking results to evaluate end-to-end performance.
\section{Experiments}
\label{sec:experiments}
\subsection{Dendritic spine detection in 2D}
We evaluate the detection performance of our method, as well as that of SpineS \cite{arguncsah2022interactive}, DeepD3 \cite{fernholz2024deepd3}, and the approach from Vogel et al. \cite{vogel2023utilizing}, using their publicly available, pre-trained models without additional fine-tuning. For SpineS and DeepD3, whose outputs are not natively bounding boxes, a conversion of the prediction is done (see Supplement) to enable a fair comparison.

\begin{figure}[hbt!]
  \centering
  \includegraphics[width=0.45\textwidth]{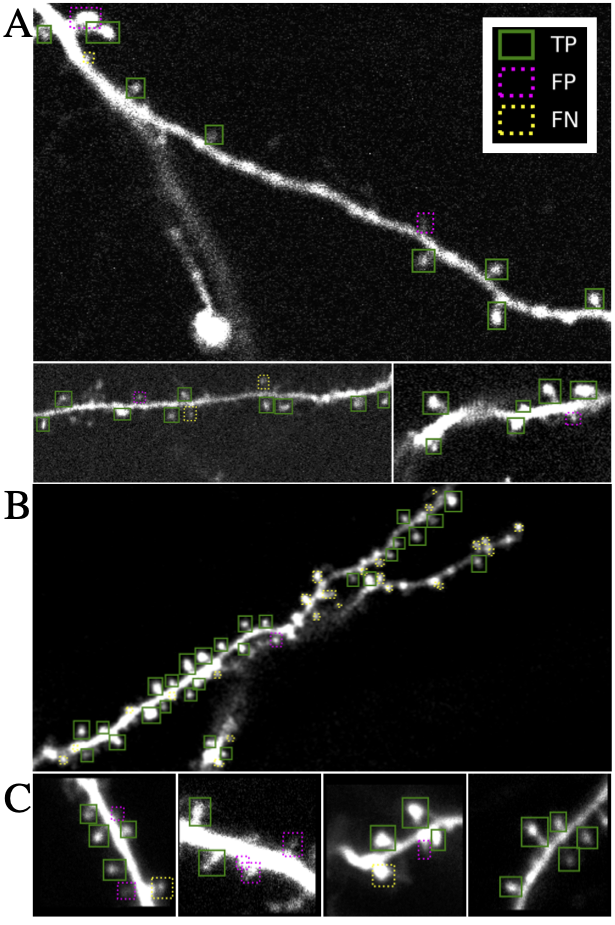}
  \caption{\textbf{Qualitative results of spine detection with SynapFlow (trained on S3D-train).} Results on representative images from the S3D dataset (\textbf{A}), the DeepD3 dataset (\textbf{B}), and the Smirnov-Garrett dataset (\textbf{C}). True positive predictions are shown in green (solid line), false positives in magenta (dotted line), and false negatives in yellow (dotted line). For visualization purposes, image contrast is enhanced.} 
  \label{fig: qualitative_results_detection}
\end{figure}

We report detection results on three datasets: our S3D dataset, the DeepD3 dataset, and the Smirnov-Garrett dataset (see Tables \ref{tab:detection-scores-S3D}-\ref{tab:detection-scores-Tavita}). Representative examples from these sets are shown in Fig.  \ref{fig: qualitative_results_detection}, highlighting some qualitative differences regarding spine density and background structures (e.g. axons). For the S3D dataset, we measure performance variability by creating four random data partitions. For each partition we train a Deformable DETR network on the training set and evaluate its performance on the corresponding test set.

A cost matrix is built using pairwise IoM scores between ground-truth boxes and predictions.  Optimal matching is then computed via the Hungarian algorithm, discarding matches with an IoM lower than 0.7 (empirically set).

We observe that the DeepD3 model consistently achieves a high recall across all three datasets. However, it also tends to over-predict dendritic spines, producing roughly twice as many detections as the ground-truth in both the S3D dataset and the Smirnov-Garrett dataset. One possible explanation is that the DeepD3 model was trained on data where manual annotations were made under the assumption that all blob-like structures were spines, as no axons nor artifacts are observable in those data. While reasonable for their data, this assumption does not hold in our dataset, which may account for the lower precision observed for both the DeepD3 model and SpineS on our S3D data. In contrast, our model and that of Vogel et al. exhibit a more conservative detection strategy. Although they generally yield lower recall than DeepD3 model and SpineS, their predictions tend to be more precise. As shown in Fig. \ref{fig: qualitative_results_detection}B, the spines missed by our model in the DeepD3 dataset are often those located on top of dendrites, likely oriented orthogonally to the imaging axis.

While the R-CNN model from Vogel et al. has the closest performance to our model on the S3D test set, its performance on the DeepD3 dataset is much lower compared to our Deformable DETR. The images for the DeepD3 data (Fig. \ref{fig: qualitative_results_detection}B) show a smaller resolution and size of spines compared to the overall image. The Deformable DETR network has shown advantages in detecting smaller objects \cite{zhu2020deformable}, which is a potential explanation for its competitive performance on the DeepD3 dataset. Another reason could be that CNNs are more prone to overfitting to the training data compared to attention-based architectures such as the Deformable DETR \cite{takahashi2024comparison}.

Although no single detection model consistently outperforms the others across all three datasets, our method consistently ranks among the top two in terms of the F1-score (Tables \ref{tab:detection-scores-S3D}-\ref{tab:detection-scores-Tavita}). This suggests that it achieves a strong balance between precision and recall, demonstrating robustness across diverse imaging samples and datasets. 

\begin{table}[hbt!]
  \centering
  \begin{center}
  \resizebox{\columnwidth}{!}{%
  \begin{tabular}{lccc} 
    \toprule
     Method & Precision  & Recall & F1-score\\
    \midrule
    SpineS  & $0.26\pm0.09$  & $0.53\pm0.03$  & $0.34\pm0.08$\\
    DeepD3   & $0.26\pm0.05$  & $\mathbf{0.73\pm0.13}$  & $0.39\pm0.07$\\
    Vogel et al. & $\underline{0.68\pm0.07}$ & $\underline{0.69\pm0.06}$ & $\underline{0.68\pm0.04}$\\
    \hline
    SynapFlow & $\mathbf{0.82\pm0.04}$ & $\mathbf{0.73\pm0.06}$  & $\mathbf{0.77\pm0.03}$\\ 
  \bottomrule
  \end{tabular}
  }
  \end{center}
  \caption{\textbf{Quantitative evaluation of 2D spine detection methods on the S3D dataset (test partition).} We report the resulting precision, recall, and F1-score. For methods that do not originally produce bounding boxes (e.g. DeepD3, SpineS), we apply a conversion procedure to approximate bounding boxes (see Supplement), enabling a consistent evaluation across methods. The best score is shown in bold, while the second best is underlined. Results are reported as mean $\pm$ standard deviation across four test sets. In the case of SynapFlow, the model is trained separately for each test set, using a training set non-overlapping with the test set.}
  \label{tab:detection-scores-S3D}
\end{table}

\begin{table}[hbt!]
  \centering
  \begin{center}
  \resizebox{\columnwidth}{!}{%
  \begin{tabular}{lcccc}
    \toprule
     Method & Precision  & Recall & F1-score & \#Detections \\
    \midrule
    Ground-truth & - & - & - & $2335$ \\
    SpineS  & $0.23$  & $\underline{0.57}$  & $0.33$ & $963$ \\
    DeepD3   & $0.77$  & $\mathbf{0.68}$  & $\mathbf{0.72}$ & $2057$ \\
    Vogel et al.  & $\mathbf{0.96}$ & $0.25$ & $0.40$ & $2335$ \\
    \hline
    SynapFlow & $\underline{0.86}$ & $0.48$  & $\underline{0.62}$ & $1297$ \\
  \bottomrule
  \end{tabular}%
  }
  \end{center}
  \caption{\textbf{Quantitative evaluation of 2D spine detection on the DeepD3 dataset \cite{fernholz2023data, fernholz2024deepd3}.} As for Table \ref{tab:detection-scores-S3D}.}
  \label{tab:detection-scores-DeepD3}
\end{table} 

\begin{table}[hbt!]
  \centering
    \begin{center}
  \resizebox{\columnwidth}{!}{%
  \begin{tabular}{lcccc} 
    \toprule
     Method & Precision & Recall & F1-score & \#Detections\\
    \midrule
    Ground-truth & - & - & - & $6928$ \\
    SpineS  & $\mathbf{0.84}$  & $\underline{0.86}$  & $\mathbf{0.82}$ & $6631$ \\
    DeepD3   & $0.52$  & $\mathbf{0.90}$  & $0.66$ & $11942$ \\
    Vogel et al. & $\underline{0.60}$ & $0.31$ & $0.41$ & $3601$ \\ 
    \hline
    SynapFlow & $0.59$ & $0.79$  & $\underline{0.68}$ & $9226$ \\
  \bottomrule
  \end{tabular}%
  }
  \end{center}
  \caption{\textbf{Quantitative evaluation of 2D spine detection on the Smirnov-Garrett dataset \cite{smirnov2018open, tavita2018labeled}.} As for Table \ref{tab:detection-scores-S3D}.}
  \label{tab:detection-scores-Tavita}
\end{table}

\subsection{Spine tracking across depth}
To improve the spatial consistency cost, slices are first aligned pairwise across depth with a rigid transformation using the ITK library \cite{mccormick_itk2014}.

To specifically evaluate depth-tracking performance, we use the manual 2D detections from the S3D dataset (test partition) as input to various depth-tracking methods and assess their output using standard MOT metrics (see Table \ref{tab:depth-tracking-scores}). We test two configurations of our approach: one that relies solely on spatial consistency (SynapFlow$_\text{sp}$), and another that incorporates spatial and appearance features (SynapFlow$_\text{sp+\text{app}}$). As with detection, trainable components (here only SynapFlow$_\text{sp+\text{app}}$) are trained four times, one for each data split, and evaluated on their corresponding test partition to capture performance variability. 

Overall, all tested methods are able to integrate 2D detections into 3D objects reasonable well. However, the IDF1 metric is systematically lower across methods, suggesting that avoiding identity switches and fragmenting trajectories remains challenging and is a potential avenue for improvement. 

Both SynapFlow variants outperform the baseline methods in terms of HOTA and AssA, highlighting the strength of our approach to accurately reconstruct 3D spine identities. Interestingly, adding an appearance-based cost (SynapFlow$_\text{sp+\text{app}}$) leads to a slight decrease in performance compared to using spatial cues alone. This may reflect that, in our dataset, spatial continuity across depth is already highly informative, and visual cues can occasionally introduce noise, particularly in regions with overlapping or visually similar spines.

\begin{figure}[hbt!]
  \centering
  \includegraphics[width=0.49\textwidth]{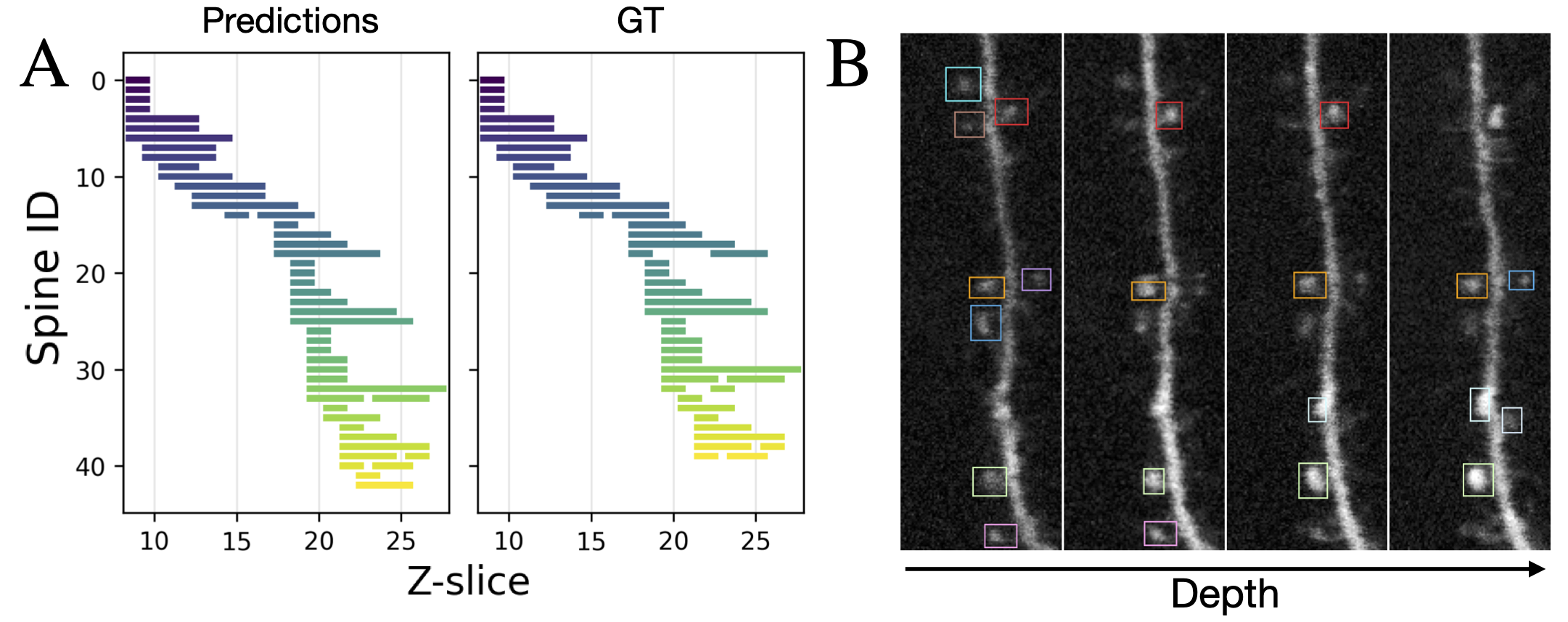}
  \caption{\textbf{Dendritic spines tracked accross depth.} \textbf{A}: Predicted vs. ground-truth (GT) associations of 2D spine instances across depth. The y-axis shows the 3D identity assigned to the 2D instance, and the x-axis shows the z-slices at which the 2D instance is located. \textbf{B}: 2D spines tracked across depth. 3D spine identity is color-coded.} 
  \label{fig: samples-depth-tracking}
\end{figure}

As shown in Fig. \ref{fig: samples-depth-tracking}A, a common failure mode of our depth-tracking method occurs when 2D spine instances are missing for several consecutive z-slices and then reappear. This often happens due to spine overlap and occlusion, and is in line with the observed drop in IDF1 performance.

\begin{table}[hbt!]
  \centering
  \begin{center}
    \resizebox{\columnwidth}{!}{%
  \begin{tabular}{lccc} 
    \toprule
     Method & HOTA $\uparrow$ & IDF1 $\uparrow$ & AssA $\uparrow$\\
    \midrule
    3D Connected Comp.  & $90.43\pm1.57$  & $85.48\pm5.21$  & $94.42\pm0.77$\\
    IoM$_{0.5}$  & $98.1\pm0.34$ & $\underline{92.16\pm2.32}$ & $96.79\pm0.60$ \\
    \midrule
    SynapFlow$_{\text{sp}+\text{app}}$ & $\underline{98.22\pm0.40}$ & $92.06\pm2.26$ & $\underline{97.08\pm0.28}$\\
    SynapFlow$_{\text{sp}}$ & $\mathbf{98.38\pm0.52}$ & $\mathbf{93.30\pm2.58}$ & $\mathbf{97.42\pm0.57}$\\ 
  \bottomrule
  \end{tabular}%
  }
  \end{center}
  \caption{\textbf{Quantitative evaluation of spine depth-tracking methods using the S3D dataset.} 2D ground-truth detections from each of the four S3D test sets are input to different depth-tracking methods, and we report here the results for standard MOT metrics. The 3D connected components algorithm was used as in \cite{fernholz2024deepd3}, and the IoM-based metric was used as in \cite{vogel2023utilizing}. Our SynapFlow$_\text{sp}$ approach relies solely on spatial information and does not involve training, while our SynapFlow$_\text{sp+app}$ approach additionally considers visual information via a Siamese network (Section \ref{subsec: method-depth-tracking}) and trained on the S3D dataset (train partition). The best score is shown in bold, while the second best is underlined. Results are reported as a mean $\pm$ standard deviation across four test sets.   In the case of SynapFlow$_\text{sp+app}$, the model is trained separately for each test set, using a training set non-overlapping with the test set.}
  \label{tab:depth-tracking-scores}
\end{table}

For the combined task of detecting and depth-tracking, we compare the performance  of our method against Trackformer, both trained on the S3D dataset. The detection module of SynapFlow used here is the Deformable DETR, as described in Section \ref{subsec: method-detection}, and the depth-tracking module is based on our weighted spatial and appearance score (Eq. \ref{eq: cost_depth_tracking}). We compute MOTA, HOTA, IDF1, and AssA for both models, and find SynapFlow to achieve better results for all these metrics (see Table \ref{tab:depth-tracking-scores-defdetr-dets}). The gap is highest for the MOTA score, which could possibly be explained by biases due to features previously learned by Trackformer on the MOT17 dataset. Nonetheless, the scores for metrics that rely less on the detection performance (IDF1 and AssA) are also higher when jointly using the detection and depth-tracking module of SynapFlow, pointing to a more accurate identification of 3D spine objects overall. 

\begin{table}[hbt!]
  \centering
  \begin{center}
    \resizebox{\columnwidth}{!}{%
  \begin{tabular}{lcccccc} 
    \toprule
     Method  & MOTA $\uparrow$ & HOTA $\uparrow$  & IDF1 $\uparrow$ & AssA $\uparrow$ & \#Dets. & \#Tracks  \\
    \midrule
    Ground-truth & - & - & - & - & $883$ & $195$\\
    Trackformer  & $46.54$ & $69.58$ & $68.03$ & $77.34$ & $1234$ & $255$ \\
    \hline
    SynapFlow$_{\text{sp}+\text{app}}$ (with DefDETR) & $\underline{61.60}$ & $\underline{73.58}$ & $\underline{79.13}$ & $\underline{80.17}$ & $881$ & $203$\\
    SynapFlow$_{\text{sp}}$ (with DefDETR) & $\mathbf{61.72}$ & $\mathbf{73.63}$ & $\mathbf{79.25}$ & $\mathbf{80.32}$ & $881$ & $200$\\
  \bottomrule
  \end{tabular}%
  }
  \end{center}
  \caption{\textbf{Quantitative evaluation of joint spine detection and depth-tracking on the S3D dataset.} Performance of spine detection and depth-tracking according to standard MOT metrics. All methods are trained on the S3D dataset (train partition). \#Dets. refers to the number of detections produced by each given method, while \#Tracks refers the number of 3D objects. The best score is shown in bold, while the second best in underlined.}
  \label{tab:depth-tracking-scores-defdetr-dets}
\end{table}

\subsection{Spine tracking across time}
As part of this work, we release the first large-scale dataset of dendritic spines manually tracked across imaging days, referred to as S2D+T (see Section \ref{subsec:dataset}), enabling the evaluation of automated methods for the tracking of spines across time. Here, we use  the S2D+T dataset to evaluate our proposed time-tracking method. By characterizing each 3D spine object using a 2D detection, we formulate the task as a MOT problem across time, as outlined in Section \ref{subsec:method-performance-eval}. We report the resulting performance scores in Table \ref{tab:time-tracking-scores}. 

\begin{figure}[hbt!]
  \centering
  \includegraphics[width=0.49\textwidth]{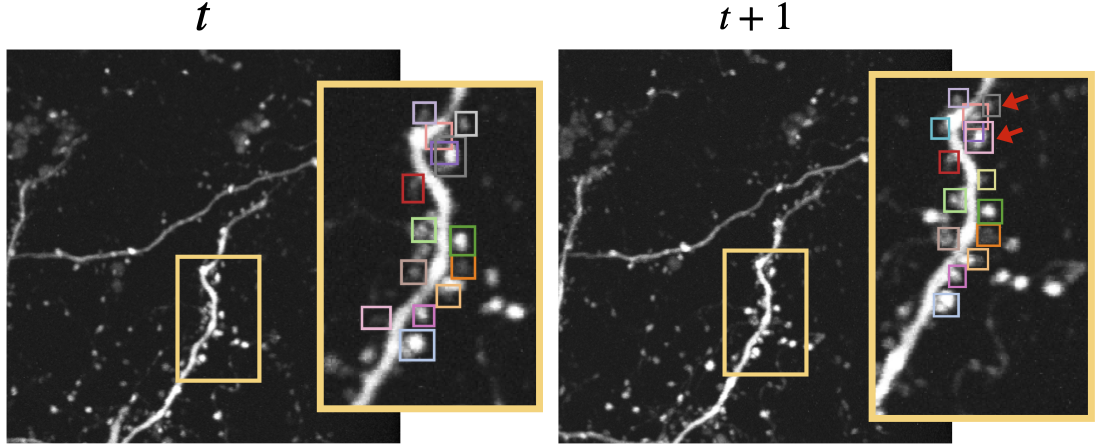}
  \caption{\textbf{Dendritic spines tracked across time.} Median spine boxes of the 3D objects are displayed here on top of the maximum intensity projection of the volume for visualization purposes. Identities over time are color-coded. Incorrect ID assignments are shown with red arrows.}
  \label{fig: qualitative-results-time-tracking}
\end{figure}

Most tracking errors occur in regions with overlapping spines, as indicated by red arrows in Fig. \ref{fig: qualitative-results-time-tracking}. In these densely populated areas, spines partially occlude each other, making it difficult to clearly delineate individual objects. This occlusion complicates both their spatial- and appearance-based identification, leading to tracking ambiguities.

\begin{table}[hbt!]
  \centering
  \begin{center}
    \resizebox{\columnwidth}{!}{%
  \begin{tabular}{lcccc} 
    \toprule
     Method  & HOTA $\uparrow$  & IDF1 $\uparrow$ & AssA $\uparrow$ & \#Tracks  \\
    \midrule
    Ground-truth & - & - & - & $1450$ \\
    SynapFlow$_{\text{sp}+\text{app}}$ & $88.37$ & $82.41$ & $78.46$ & $1640$\\
    SynapFlow$_{\text{sp}}$ & $\mathbf{90.23}$ & $\mathbf{84.54}$ & $\mathbf{81.78}$ & $1529$ \\
  \bottomrule
  \end{tabular}%
  }
  \end{center}
  \caption{\textbf{Quantitative evaluation of the proposed spine time-tracking method using the dataset S2D+T.} The S2D+T dataset is processed by two variants of the proposed time-tracking method. The performance is assessed using standard MOT metrics. SynapFlow$_{\text{sp}+\text{app}}$ represents the variant that employs both a spatial and an appearance cost, while SynapFlow$_\text{sp}$ performs the matching across time only based on spatial consistency. Tracking is performed across four timepoints distributed over two imaging days separated by 3 days. The number of objects tracked across time is shown as \#Tracks. The best score is shown in bold.}
  \label{tab:time-tracking-scores}
\end{table}

\subsection{Computation of spine features}
\begin{figure}[hbt!]
  \centering
  \includegraphics[width=0.43\textwidth]{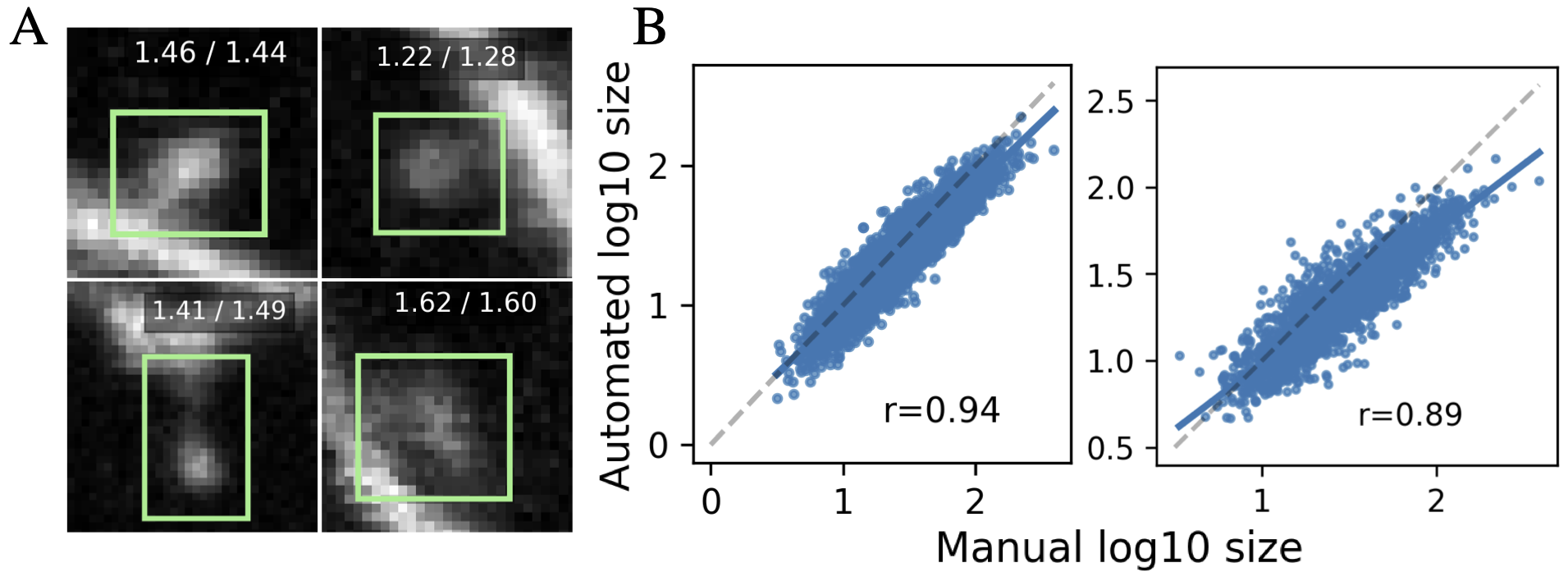}
  \caption{\textbf{Automated spine size estimation.} \textbf{A}: Dendritic spines from the ground-truth detections with the automated vs. manually estimated size. \textbf{B}: Pearson correlation of manual spine sizes (in arbitrary units) against the sizes estimated automatically, either given the ground-truth detections, $\text{n}=3966$ (left), or from the 3D predicted objects, $\text{n}=2242$ (right).}  
  \label{fig: size-computation}
\end{figure}

To evaluate the accuracy of our automated spine size estimator in isolation, we use the polygon annotations from dataset S2D+T, approximate them with bounding boxes, and automatically estimate their size. We then compare these estimates to the corresponding manual measurements using Pearson correlation (Fig. \ref{fig: size-computation}A, B left), and observe a strong agreement, indicating that our method provides reliable size estimations. Notably, we find a slight underestimation for larger spine sizes. Despite the potential inaccuracies introduced by the use of bounding boxes instead of segmentation masks, the estimated sizes remain remarkably consistent with the manual annotations.

\begin{figure}[hbt!]
  \centering
  \includegraphics[width=0.4\textwidth]{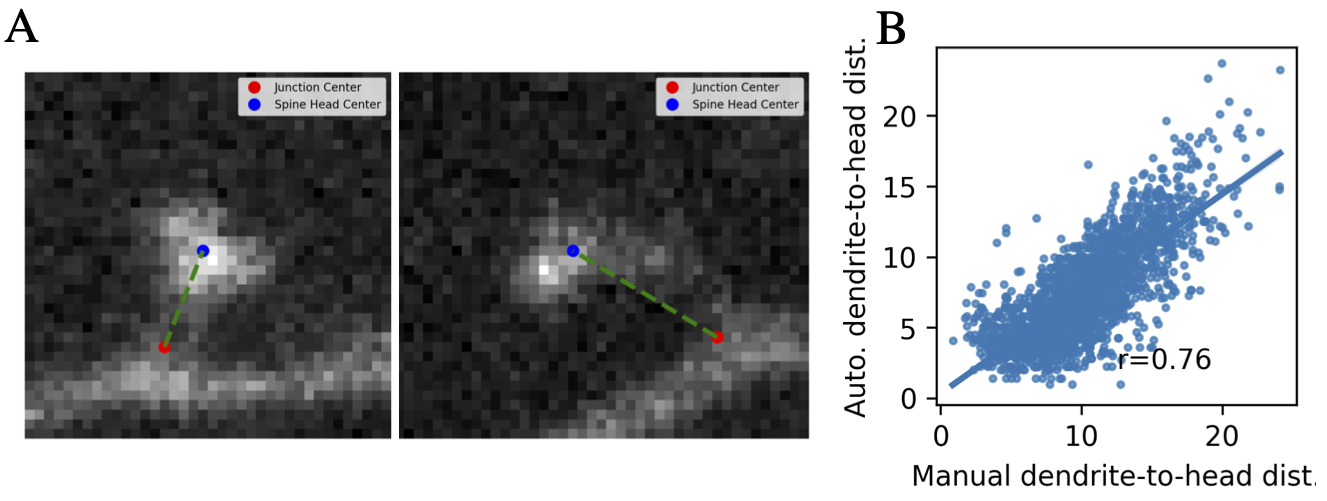}
  \caption{\textbf{Automated spine-to-dendrite distance estimation.} \textbf{A}: Spines with their automatically estimated spine-to-dendrite distance. \textbf{B}: Pearson correlation between manually drawn and automatically computed spine-to-dendrite distance for the automatically detected 3D spine objects ($\text{n}=2242$).}  
  \label{fig: spine-to-dend-computation}
\end{figure}

Furthermore, to assess the accuracy of the size estimates in the presence of potential discrepancies introduced by the automated detection and depth-tracking steps, we also compute the sizes of the predicted 3D spine objects and compare them to the manually estimated sizes. The correspondence between predictions and ground-truth detections is established via bijective matching with the Hungarian algorithm and the resulting pairs of size measurements are evaluated. The resulting correlation is slightly lower than that obtained using ground-truth detections, but remains fairly high (Fig \ref{fig: size-computation}B, right).  An analogous evaluation was performed for the spine-to-dendrite distance computation, showing that our estimates closely match manual annotations (Fig. \ref{fig: spine-to-dend-computation}A).

\section{Discussion}
\label{sec:discussion}
We introduce SynapFlow, a modular pipeline to detect, depth-track, and time-track dendritic spines, while also estimating their size and spine-to-dendrite distance. The modularity of SynapFlow enables flexible adaptation of each component, which can be individually refined or replaced as methods and data evolve, supporting continual improvement. We demonstrate the robustness of the detection unit by evaluating it not only across four independent test partitions of our own dataset but also on publicly available spine datasets. The results confirm that SynapFlow maintains solid performance across different imaging conditions and data sources. In addition to detection, we validate the depth-tracking and time-tracking modules using two new datasets released alongside this work. In particular, we here introduce, to the best of our knowledge, the first large-scale longitudinal imaging dataset with spines manually tracked across time, filling an important gap in the field and supporting systematic evaluation of temporal spine dynamics.

Some limitations remain. We could not quantitatively assess the effect of the depth-consistency term in time-tracking due to the lack of depth-annotated time series, though visual inspection indicates it is crucial to prevent matching spines located far in depth. Moreover, both manual and automated detection omits spines oriented orthogonally to the imaging axis, which can lead to apparent turnover events.

Regarding appearance-based matching, our experiments show no improvement when adding visual cues, suggesting spatial information to be often dominant in our data. Still, training the CNN encoder on cases where spatial cues fail could strengthen its contribution. An attention-based architecture combining spatial and visual features directly might help resolve ambiguity. 

The variability observed in detection performance across datasets highlights the need for more comprehensive, diverse, and standardized benchmarks. Also, extracting further morphological features, such as neck length, head roundness, or volume will require richer annotations but could provide further biological insights.

We acknowledge the importance of the accessibility and user-friendliness of the approach, and are committed to working towards this goal throughout future work.

By sharing our pipeline and datasets, we hope SynapFlow to provide a solid baseline for advancing the automated, large-scale, and objective analysis of dendritic spine dynamics.
\section{Acknowledgements}
The research leading to these results has received funding from the Deutsche Forschungsgemeinschaft (DFG, German Research Foundation) SPP 2041 "Computational Connectomics" (JT, SR, MK); the DFG 414985841 GRK 2566 "iMOL" (MK); the DFG Research Unit FOR 5368 ARENA (MK, JT), the DFG DIP ‘Neurobiology of Forgetting’ (SR, MK); Germany’s Excellence Strategy EXC 3066/1 “The Adaptive Mind”, Project No. 533717223 (JT); the Speyer'sche Hochschulstiftung (MK); and the Johanna Quandt foundation (JT).

\clearpage


{
    \small
    \bibliographystyle{ieeenat_fullname}
    \bibliography{main}
}

\end{document}